\documentclass[twoside,11pt]{article} 
\usepackage{jmlr2e} 

\usepackage[utf8]{inputenc}
\usepackage[T1]{fontenc}
\usepackage{amsmath}
\usepackage{amsfonts}
\usepackage{tabularx}
\usepackage{booktabs}
\usepackage[ruled,vlined]{algorithm2e}
\usepackage{graphicx}
\usepackage{verbatim}   
\usepackage{framed}     
\usepackage{tikz}
\usetikzlibrary{shapes,arrows}
\usepackage{tipa}
\usepackage{xspace}
\usepackage{arydshln}
\usepackage{float} 
\usepackage{hyperref}

\newcommand{\argmax}{\operatornamewithlimits{argmax}}
\newcommand{\nbx}{\emph{NaiBX}\xspace}
\newcommand{\tm}{$NBX_{True M}$\xspace}

\newcommand{\nbc}{\emph{CC$_{NBC}$}\xspace}
\newcommand{\enbc}{\emph{ECC$_{NBC}$}\xspace}
\newcommand{\svm}{\emph{CC$_{SVM}$}\xspace}
\newcommand{\esvm}{\emph{ECC$_{SVM}$}\xspace}
\newcommand{\rk}{\emph{{RAkEL}}\xspace}
\newcommand{\mline}{\cdashline{2-11}[0.3pt/2pt]}

\title{Naive Bayes Classification for Subset Selection}

\editor{}
    
\author{\name Luca Mossina \email luca.mossina@isae-supaero.fr\\
    \name Emmanuel Rachelson \email emmanuel.rachelson@isae-supaero.fr \\
    \addr  
    Département d'Ingénierie des Systèmes Complexes (DISC)\\
    ISAE-SUPAERO - Institut Supérieur de l'Aéronautique et de l'Espace\\
    10 avenue Edouard Belin - BP 54032 - 31055 TOULOUSE CEDEX 4 FRANCE}

\ShortHeadings{Naive Bayes Classification for Subset Selection}{Mossina and Rachelson}

\begin{document}
\maketitle

\begin{abstract}%
This article focuses on the question of learning how to automatically select a subset of items among a bigger set. We introduce a methodology for the inference of ensembles of discrete values, based on the Naive Bayes assumption. Our motivation stems from practical use cases where one wishes to predict an unordered set of (possibly interdependent) values from a set of observed features. This problem can be considered in the context of Multi-label Classification (MLC) where such values are seen as labels associated to continuous or discrete features. We introduce the \nbx algorithm, an extension of Naive Bayes classification into the multi-label domain, discuss its properties and evaluate our approach on real-world problems.
\end{abstract}


\section{Introduction}
Consider the problem of mapping a set of features to a set of labels. For instance, one may wish to associate the cooking preferences of tonight's guests to a list of ingredients for a cake, or map an image to a set of animal species present in the image, or again a text document to a list of topics. In the cake toy-example, it is reasonable to assume that, for the shopping part, the order in which the ingredients are presented is irrelevant. However, it is important that all ingredients for the cake are on the shopping list rather than a mix of ingredients between two probably acceptable cakes. Moreover, if the guests prefer a certain type of cake and one only needs brown sugar for its recipe, but could probably substitute it with white sugar, it is a desirable property of the cooking assistant that it does not predict both brown and white sugar with high probability in the shopping list. Such intuitive statements are key features of multi-label classification, when predicted labels are (conditionally) dependent on the presence of other labels, and are issues we attempt to address in this paper.

This article deals with such problems where one wishes to predict a set of discrete-valued variables with no specific order and where the target variables are interdependent. Where  multi-class classification maps an example $\mathbf{x}$ to a single element within a (possibly large) set of classes, multi-label classification maps each $\mathbf{x}$ to a \emph{subset} of all possible labels within the class set.

Practical examples range from document labeling to gene functional classification.
One particular new application, which motivates the current work, deals with the prediction of which electricity power plants need to have their production schedule changed when unexpected events occur in the network, in order to quickly re-optimize the production so as to answer the power demand.

The problem of predicting a subset $y \subset \mathcal{Y}$ of labels given a set of features $\mathbf{x}$ can be seen as one of multi-class classification if we consider a target class as being one of the possible subsets in the powerset $ \mathcal{P(Y)}$. As one deals with a powerset, this number might be very large as it grows exponentially with the size of the labels set. To counter this, we first learn to predict the correct subset size $m$ for a given $\mathbf{x}$, then predict a first value given $\mathbf{x}$ and $m$, then a second value given $\mathbf{x}$, $m$ and the first value, and so on until we reach the predefined size of the subset we wish to predict. This allows us to retain the conditional dependencies between values within the set. For this purpose, we construct a cascade of predictors (where predictor number $k$ predicts the $k$-th element in the subset) and we suppose that each of these predictors is a Naive Bayes Classifier (NBC). This hypothesis induces a dramatic simplification of the predictors computation's complexity (both in time and space), and the overall learning task boils down to training the elementary parameters of a single NBC that can be used for prediction of every element in the subset. Our presentation unfolds as follows. We recall the principle and properties of Naive Bayes classification in Section \ref{sec:nbc}. Then, in Section \ref{sec:prediction}, we review the question of multi-label classification, formalize the problem of learning a probabilistic classifier for a subset of labels, introduce a general ``cascade of predictors'' method in which we finally apply Naive Bayes Classifiers. We derive a learning algorithm called \nbx and discuss its properties in Section \ref{sec:naibx}. Experimental validation results and comparisons are presented in Section \ref{sec:experiments}. Finally, we summarize and conclude in Section \ref{sec:conclusion}.

\section{Naive Bayes Classification}
\label{sec:nbc}

%

Classification, a key problem in supervised learning, consists in mapping a tuple of features $\mathbf{x} = (x_1,\ldots,x_n)$ belonging to a feature space $\mathcal{X}$, to a class $c$ in the set $\mathcal{C}$ of possible targets. Different approaches to classification have been discussed in the Machine Learning literature; for a review, see \citet{kotsiantis2007}, or the fundamental textbook by \citet{hastie2001}.

Let $\mathbf{X}=(X_1,\ldots,X_n)$ be the random vector of observations $\mathbf{x}$, taking its values in $\mathcal{X}$, and let $C$ be the random variable describing the class associated to $\mathbf{X}$. A probabilistic classifier $f$ assigns the class $c$ to a new observation $\mathbf{x}$ if $c$ maximizes the conditional probability of $C=c$ given that $\mathbf{X}=\mathbf{x}$. More formally

    \begin{equation}
        f(\mathbf{x}) = \argmax_{c\in\mathcal{C}} \mathbb{P}(C=c|\mathbf{X}=\mathbf{x}).
    \label{eq:eq1}
    \end{equation}

According to Bayes rule, this probability can be written as
$$\mathbb{P}(C=c|\mathbf{X}=\mathbf{x}) = \frac{\mathbb{P}(C=c)\mathbb{P}(\mathbf{X}=\mathbf{x}|C=c)}
        {\mathbb{P}(\mathbf{X}=\mathbf{x})}.$$

Consequently, evaluating the argument under the $\max$ operator in Equation \ref{eq:eq1} boils down to evaluating the probability $\mathbb{P}(C=c)$ of observing class $c$ and the probability $\mathbb{P}(\mathbf{X}=\mathbf{x}|C=c)$ of observing a sample $\mathbf{x}$ among those of class $c$. This last quantity is a joint distribution over $n=\dim(\mathcal{X})$ variables which might be difficult to estimate. The Naive Bayes assumption consists in supposing that all features $X_i$ of $\mathbf{X}$ are conditionally independent given the value of the class variable; that is $\forall (i,j) \in [1,n]^2, \quad \mathbb{P}(X_i|C,X_j) = \mathbb{P}(X_i|C)$. This implies that
$$\mathbb{P}(C=c|\mathbf{X}=\mathbf{x}) = \frac{\mathbb{P}(C=c)\prod\limits_{i=1}^n\mathbb{P}(X_i=x_i|C=c)}{\mathbb{P}(\mathbf{X}=\mathbf{x})},$$

and eventually,
\begin{equation}
f(\mathbf{x}) = \argmax_{c\in\mathcal{C}}\left[ \mathbb{P}(C=c)\prod\limits_{i=1}^n\mathbb{P}(X_i=x_i|C=c)\right].
\label{eq:nbc}
\end{equation}

Thus, constructing a Naive Bayes classifier consists in estimating values for $\mathbb{P}(C)$ and $\mathbb{P}(X_i|C)$, given a set of training examples $\left\{\left(\mathbf{x}^l, c^l)\right)\right\}_{l\in[1,L]}$, so as to solve the maximization problem of Equation \ref{eq:nbc} given a new observation $\mathbf{x}_{new}$.

\subsection{Continuous Features in NBCs}

In the case of continuous $X_i$ features, one generally assumes a parametric probability density function $p(X_i=x_i|C=c)$ for the distribution of $X_i$ given $C$. In this case, Equation \ref{eq:nbc} needs to be rewritten as
\begin{equation}
f(\mathbf{x}) = \argmax_{c\in\mathcal{C}}\left[ \mathbb{P}(C=c)\frac{\prod\limits_{i=1}^n p(X_i=x_i|C=c)}{p(\mathbf{X}=\mathbf{x})}\right],
\label{eq:nbc-cont}
\end{equation}
and the denominator in Equation \ref{eq:nbc-cont} is computed as
\begin{equation*}
p\left(\mathbf{X}=\mathbf{x}\right) = \sum\limits_{c\in\mathcal{C}} p\left(\mathbf{X}=\mathbf{x}|C=c\right).
\end{equation*}

The $p\left(\mathbf{X}=\mathbf{x}|C=c\right)$ elements in the last expression can be decomposed further into products of univariate distributions using again the Naive Bayes assumption.

\subsection{Optimality of NBCs}

Although the Naive Bayes assumption is rarely true in most real-world applications, NBCs perform surprisingly well and are often competitive against more sophisticated methods (such as Classification Trees, Support Vector Machines or Neural Networks for instance). The conditions for the optimality of NBCs given the conditional independence assumption (hardly ever met, in practice) have been studied for instance in \citep{domingos1997} and, more recently, in \citep{zhang2004, zhang2005}. Intuitively, since the conditional independence assumption almost never holds, the probability estimation perfomance of NBCs may be poor. The authors of \citep{zhang2004} demonstrate it is the distribution of dependencies among attributes $X_i$ that affects the classification performance, rather than the dependencies themselves. They derive the conditions for optimality of NBCs under which it is shown that the class having maximum estimated probability remains the correct one, despite the fact that those estimations are flawed.

%

\section{Multi-label Classification}
\label{sec:prediction}

The canonical classification task \citep{hastie2001} in machine learning deals with binary targets: given a vector of observations $\mathbf{x}$, one wants to associate to it the most appropriate class $c$ in $\mathcal{C}$, where $|\mathcal{C}|=2$. If we extend $\mathcal{C}$ to be a set of any finite amount of target classes we are then facing a multi-class problem: we must choose one option among many available. Binary methods such as Support Vector Machines can be extended \citep{crammer2002} to handle such cases via the construction of several \emph{one-against-all} classifiers at the cost of storing more parameters, while others such as NBC or decision trees natively support multi-class problems.

In the present paper we address the more general problem of multi-label classification (MLC). Given a set $\mathcal{C}$ of options (``labels''), we look for the most appropriate subset of those options, denoted by 
        $$f_{classifier} : \mathcal{X} \longrightarrow \mathcal{P}(\mathcal{C}),$$
where the number of labels to assign to each observation is itself part of the problem.

Often one performs a \emph{problem transformation}, that is, algorithms for binary and multi-class classification are adapted to handle multi-label problems.
One such transformation is known as the \emph{Label Powerset} (LP) method \citep{tsoumakas07,tsoumakas10}, that consists in turning a multi-label problem into a multiclass one. This operates by mapping $\mathbf{x} \in \mathcal{X}$ to one of the sets $\mathbf{c}_k = \{c_j\}$ that compose $\mathcal{P}(\mathcal{C})$, in practice usually a big set. While LP methods offer the advantage of taking into account label dependencies \citep{viola01}, as the number of classes scales exponentially with the number of labels, this approach cannot cope with rich sets of labels. To counter this, \citet{Tsoumakas2011} developed the RAkEL algorithm (\textit{RAndom k-LabEL sets}), a variation on LP that trains a series of $m$ models whose targest are a subset of $k$ labels from those available.

Another approach consists in independently training a binary classifier for each admissible label $c_j\in \mathcal{C}$ thus obtaining as many models as there are labels; a method known as \emph{Binary Relevance} (BR) \citep{tsoumakas07, read2009}. It is worth noting, however, that such an approach implicitly assumes conditional independence between labels since it predicts them independently.

\subsection{Classifiers Chains}
\label{sec:subCC}

In MLC, target vectors can be represented in two equivalent ways. Assuming for example that the set of available labels is $\mathcal{Y} = \{1,2,...,10\}$, then one can seamlessly express the targets in the following alternative fashions:

    \begin{itemize}
        \item $\textbf{y} = \texttt{[- - 3 - 5 6 7 - 9 -]}$,
        \item $\textbf{y} = \texttt{[0 0 1 0 1 1 1 0 1 0]}$.
    \end{itemize}

The second one is what is commonly adopted in the literature \citep{tsoumakas10}, where models are trained to predict the presence of label $l_i$ through a binary variable $Y_i \in \{0,1\}$.

Stemming from the principles of BR, \citet{read2009} extended the method by taking into account information about label interdependence. They proceed in an incremental way, at each step taking into account what was predicted previously. Their \textit{Classifier Chains} (CC) meta-algorithm first starts by predicting whether label $l_1$ is to be included in the target vector. It then continues predicting the presence of the second label $l_2$ given the information contained in the data ($X$) \textit{and} whether $l_1$ was included in the target vector ($\hat{Y}_{1} \in \{0,1\}$). At the $i$-th step we have

    $$ \hat{Y}_i = h(X \mid \hat{Y}_{i-1}, \hat{Y}_{i-2}, ..., \hat{Y}_{1}).$$

The order of evaluation of the binary labels $Y_i \in \{0,1\}$ can affect negatively the performance of the algorithm.
Their \textit{Ensemble of Classifier Chains} (ECC) is a possible countermeasure to this limitation: they extend their methodology by training $m$ different predictors on random permutations of the labels to then operate a \textit{bagging} \citep{Breiman96} step for the selection, via a threshold function, of the best labels. So for example one could have $h^1(Y_1, Y_4, Y_3, Y_2)$, $h^2(Y_3, Y_1, Y_2, Y_4)$ and so on.
\citet{CCbayes} refined this method with their \textit{Probabilistic Classifier Chains} (PCC) at the cost of dramatically increased computational costs. While at each iteration of CC, we get the prediction of a label (either "1" or "0"), with PCC we obtain the full joint distribution of the labels. Exploiting the chain rule of probability at each step one gets

    $$ \mathbb{P}(Y_{1}, ..., Y_{L} \mid X) = f_1(X) \cdot \prod_{i=2}^{L} f_i(X, Y_1, ..., Y_L),$$

    where $ f_i(X) = \mathbb{P}(Y_1=1 \mid X, Y_{i-1}, ..., Y_{1}) $ is a probabilistic classifier predicting the probability of label $l_i$ being included in the target vector. To accomplish this, at each level of the chain they need to compute a joint distribution, which is the cause of its high computational costs.

\subsection {Complexity of NBC for multi-label classification}
To set ideas and simplify the presentation, we shall suppose that $\mathcal{X}=\mathbb{R}^n$ and that the $X_i|C$ are distributed following a Normal distribution $\mathcal{N}(\mu_{ic},\sigma_{ic})$ (although our work straightforwardly extends to more general families of variables and distributions). Consequently, given a training set $\left\{\left(\mathbf{x}^l, c^l)\right)\right\}_{l\in[1,L]}$, estimating the parameters of an NBC consists in estimating and storing $|\mathcal{C}|-1$ values to describe $\mathbb{P}(C)$ and $2n|\mathcal{C}|$ values to describe all $p(X_i|C)$.
More generally, the space requirements for estimating and storing an NBC are in $\mathcal{O}(\kappa n |\mathcal{C}|)$, where $\kappa$ is the number of parameters required to characterize a single univariate distribution $p(X_i|C=c)$.

If the class space $\mathcal{C}$ is the powerset $\mathcal{P}(\mathcal{Y})$ of a label set $\mathcal{Y}$, then the number of possible classes is $|\mathcal{C}| = 2^{|\mathcal{Y}|}$ and the space requirements of the corresponding NBC are in $\mathcal{O}(\kappa n 2^{|\mathcal{Y}|})$. These requirements quickly become impractical for real-life applications; for instance, a problem with 10 Gaussian features $X_i$ and 100 labels yields an order of magnitude for the space complexity of the corresponding NBC of $10^{31}$ values stored. This specific issue motivates the developments presented below.

\subsection{Motivating Example}
\label{mot_ex}
Before presenting the main contribution of this paper, we introduce a practical application case which motivated the present work, although detailed results for his problem are beyond the scope of this paper. In the context of the unit commitment problems, energy providers finely plan their production one day ahead, given a forecast of the demand. Often, unpredicted events (temperature variations, utility failure, etc.) affect this demand and the original plan must be adapted in order to meet the updated demand while minimizing costs for the producer.

Optimization software for electricity production planning aim at solving extremely large combinatorial optimization problems which may take a long time to solve to optimality. Since the new demand is known within a very short delay, any operation that minimizes the running time of this re-optimization allows a quick return to a balanced network and an economical benefit for the producer. One such operation consists in selecting a subset of the network's plants whose production plan should be modified, thus defining a reduced problem, whose resolution time is compatible with the operational requirements.

Let $K$ be the number of power plants in the network and let us suppose that one wishes to select $m$ of them to be re-planned. The corresponding NBC that predicts the $m$ values has space requirements in $\mathcal{O}\left(n\kappa \frac{K!}{(K-m)!}\right)$, where $n$ is the number of features characterizing the problem (e.g. the demand at different moments in the day). For a network with 150 plants and a subset of 30 plants to be re-optimized, $\frac{K!}{(K-m)!} \sim 10^{63}$. This makes the brute-force construction of a NBC practically infeasible. Moreover, if the number $m$ of re-optimized plants is not fixed beforehand, then the space complexity of the NBC is even worse.

It is important to note in this example that the dependencies between plants that participate in the re-optimization should be preserved by the classification algorithm. Suppose for instance that in the re-optimized production schedule, only the program of plant $A$ has changed. Suppose, moreover, that plant $A$ has a twin plant $B$ in the network with the exact same characteristics, then an equivalent optimal schedule can be found by switching plants $A$ and $B$. The optimal set of power-plants predicted by a perfect classification algorithm, should contain either plant $A$ or plant $B$ but not both. This aspect of \emph{conditional dependencies} among labels is a key issue to be held into consideration when working on MLC (which cannot be captured by the BR approach).

\subsection{Cascades of Predictors}

One intuitive way to tackle the problems illustrated above is to consider that selecting a given-size subset consists in choosing a first element in $\mathcal{Y}$, then a second given the first, then a third given the first and second, and so on until one reaches the appropriate subset size. In other words, selecting a subset of $\mathcal{Y}$ can be done by choosing an ordered sequence of values of $\mathcal{Y}$ if our selection function at each step effectively re-creates the correct unordered subset.
This approach differs from classifiers chains since it does not predict in sequence whether the $|\mathcal{Y}|$ labels belong or not to the target, but rather picks them incrementally. Notably, the cascade architecture does not rely on an a priori ordering of the labels (contrarily to CC).

Let $Y$ be the random variable describing the subset of $\mathcal{Y}$ that should be associated to $\mathbf{x}$. The target of the classification algorithm is to learn the correct mapping from $\mathbf{x}$ to realizations of $Y$. We write $\bar{y}$ such realizations of $Y$ to avoid confusion with vectors $\mathbf{y}$ of values of $\mathcal{Y}$. Then the classifier $f$ we are searching for is
\begin{equation}
f(\mathbf{x})=\argmax_{\bar{y}\in\mathcal{P}(\mathcal{Y})} \mathbb{P}(Y=\bar{y} | \mathbf{X}=\mathbf{x}).
\label{eq:prob_class_multi}
\end{equation}

In order to sequentially select the elements of the optimal $\bar{y}$ as stated intuitively, one would wish to decompose the probability of Equation \ref{eq:prob_class_multi} into elementary probabilities related to each element $y_i$ of $\bar{y}$. Such elementary probabilities are related to the random event $``y_i \in Y"$. Let $M$ be the random variable describing the size of $Y$. Then a subset $\bar{y}$ is composed of elements $y_1$ until $y_k$, where $M$ takes the value $k$. Given a collection of labels $y_1,\ldots,y_k$ ($\bar{y}=\{y_1,\ldots,y_k\}$) belonging to $\mathcal{Y}$, the following statements hold:
\begin{gather}
``y_1 \in Y" \land \ldots \land ``y_k \in Y" \Leftrightarrow \bar{y} \subset Y \label{eq:equiv1}\\
``y_1 \in Y" \land \ldots \land ``y_k \in Y" \land ``\textrm{all others }\not\in Y" \Leftrightarrow \bar{y} = Y \label{eq:equiv2}\\
``y_1 \in Y" \land \ldots \land ``y_k \in Y" \land ``M=k" \Leftrightarrow \bar{y} = Y \label{eq:equiv3}
\end{gather}

Equation \ref{eq:equiv1} expresses the fact that individual properties on the $y_i$ can help characterize the probability that a given subset $\bar{y}$ is included in $Y$. Equation \ref{eq:equiv2} helps expressing that Y is precisely equal to such a subset $\bar{y}$. Its formulation is equivalent to the target of CC algorithms.
Finally, Equation \ref{eq:equiv3} is of particular interest to us since it states that the subset that is both included in $Y$ and has the same size as $Y$ is precisely equal to $Y$.

We use Equation \ref{eq:equiv3} to decompose the probability estimate of the probabilistic classifier in Equation \ref{eq:prob_class_multi}, using the chain rule
\begin{multline}
\mathbb{P}(Y=\bar{y} | \mathbf{X}=\mathbf{x}) = \mathbb{P}(M=m|\mathbf{X}=\mathbf{x}) \times
\mathbb{P}(\bar{y}\subset Y|\mathbf{X}=\mathbf{x}, M=m).\label{eq:proba_m_times_y}
\end{multline}

As a notational shortcut, for any sequence of values $y_1,\ldots,y_i\in \mathcal{Y}$, we introduce
$$p(y_i|\mathbf{x},m,y_1,\ldots,y_{i-1}) =
\mathbb{P}(y_i \in Y|\mathbf{X}=\mathbf{x}, M=m, y_1\in Y, \ldots, y_{i-1}\in Y).$$

Then, using the chain rule again on the second term of the product of Equation \ref{eq:proba_m_times_y}, we can write
\begin{align*}
\mathbb{P}(Y=\bar{y} | \mathbf{X}=\mathbf{x}) = & \ p(y_m| \mathbf{x}, m, y_1, \ldots, y_{m-1}) \times \\
 & \quad p(y_{m-1}| \mathbf{x}, m, y_1, \ldots, y_{m-2}) \times \\
 & \quad\quad \ldots \times\\
 & \quad\quad\quad p(y_2 | \mathbf{x}, m, y_1) \times \\
 & \quad\quad\quad\quad p(y_1 | \mathbf{x}, m) \times \\
 & \quad\quad\quad\quad\quad \mathbb{P}(M=m | \mathbf{X}=\mathbf{x}),
\end{align*}

and more concisely
\begin{equation*}
\mathbb{P}(Y=\bar{y} | \mathbf{X}=\mathbf{x}) = \ \mathbb{P}(M=m | \mathbf{X}=\mathbf{x}) \times \prod\limits_{i=1}^m p(y_i| \mathbf{x}, m, y_1, \ldots, y_{i-1}).
\end{equation*}

So, writing $s(\mathbf{x})=\max\limits_{\bar{y}\in\mathcal{P}(\mathcal{Y})} \mathbb{P}(Y=\bar{y} | \mathbf{X}=\mathbf{x})$, Equation \ref{eq:prob_class_multi} turns into
\begin{equation*}
s(\mathbf{x}) = \max_{\bar{y}\in\mathcal{P}(\mathcal{Y})} \left( \mathbb{P}(M=m | \mathbf{X}=\mathbf{x}) \times \prod\limits_{i=1}^m p(y_i| \mathbf{x}, m, y_1, \ldots, y_{i-1}) \right).
\end{equation*}

\citet{CCbayes} analyze formally how different algorithms in the literature solve the maximization problem of Equation \ref{eq:prob_class_multi}, rephrased above using Equation \ref{eq:equiv3}. They argue that the Bayes optimal classifier solves this maximization problem to optimality. The classifier chain approach, however, exploits Equation \ref{eq:equiv2} and adopts a greedy search heuristic consisting in incrementally picking the most (marginally) probable labels in a predefined (artificial) order. Our cascade architecture somehow falls in between these two extremes. It adopts a greedy, possibly sub-optimal search method that incrementally picks labels in the label set, but does not rely on any predefined ordering of the labels.

The cascade architecture searches for a solution to the maximization problem of Equation \ref{eq:prob_class_multi} by computing the heuristic score function
\begin{align*}
s(\mathbf{x}) =& \max\limits_{y_m\in\mathcal{Y}} \Bigg[p(y_m| \mathbf{x}, m, y_1, \ldots, y_{m-1}) \times \\
 & \quad \max\limits_{y_{m-1}\in\mathcal{Y}} \bigg[p(y_{m-1}| \mathbf{x}, m, y_1, \ldots, y_{m-2}) \times \\
 & \quad\quad \ldots \times\\
 & \quad\quad\quad \max\limits_{y_{2}\in\mathcal{Y}} \Big[p(y_2 | \mathbf{x}, m, y_1) \times \\
 & \quad\quad\quad\quad \max\limits_{y_{1}\in\mathcal{Y}} \big[p(y_1 | \mathbf{x}, m) \times \\
 & \quad\quad\quad\quad\quad \max\limits_{m \in [0,|\mathcal{Y}|]} \mathbb{P}(M=m | \mathbf{X}=\mathbf{x}) \big]\Big]\bigg]\Bigg].
\end{align*}

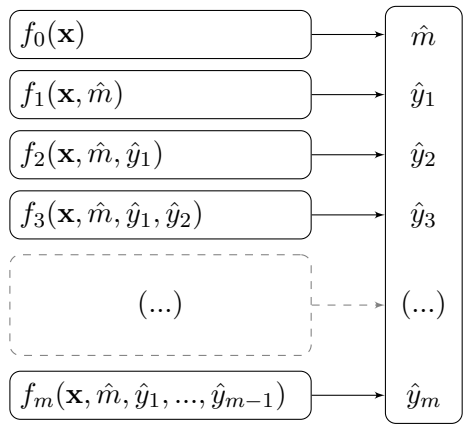
\begin{figure}
    \begin{center}
        \tikzstyle{classif} = [rectangle, draw, text width=9.8em,
                                   rounded corners, minimum height = 1.55em]

        \tikzstyle{points} 	= [rectangle, draw=gray, dashed, text width=9.8em, text centered,
                rounded corners, minimum height = 3.5em]

        \tikzstyle{targetplus} 	= [rectangle, draw, fill=none, 
        text width=2em, text centered, rounded corners, minimum height = 1.5em]

        \tikzstyle{target} 	= [rectangle, fill=none, 
	        text width=2em, text centered, rounded corners, minimum height = 1.5em]
        \tikzstyle{vert} 	= [rectangle, draw, fill=none,
	        text width=2em, text centered, rounded corners, minimum height=14.5em]
        \tikzstyle{line} 	= [draw, -latex']

        \begin{tikzpicture}[node distance = 2cm, auto]
        \node [classif] (step0) {$ f_0(\textbf{x})$};
        \node [classif, below of=step0, node distance=0.8cm] (step1) {$ f_1(\textbf{x}, \hat{m})$};
        \node [classif, below of=step1, node distance=0.8cm] (step2) {$ f_2(\textbf{x}, \hat{m}, \hat{y}_1)$};
        \node [classif, below of=step2, node distance=0.8cm] (step3) {$ f_3(\textbf{x}, \hat{m}, \hat{y}_1, \hat{y}_2)$};

        \node[points,		below of= step3, node distance=1.2cm](punkt) {$ (...) $};

        \node [classif, below of=punkt, node distance=1.2cm] (stepm) {$ f_m(\textbf{x},  \hat{m}, \hat{y}_1, ..., \hat{y}_{m-1})$};

        \node [target,	right of=step0,  node distance=3.5cm] (target0) {$ \hat{m}  $};
        \node [target,	right of=step1,  node distance=3.5cm] (target1) {$ \hat{y}_1 $};
        \node [target,  right of=step2,  node distance=3.5cm] (target2) {$ \hat{y}_2 $};
        \node [target,  right of=step3,  node distance=3.5cm] (target3) {$ \hat{y}_3 $};

        \node [vert, 	right of=step3, node distance=3.5cm] (vert02) {};

        \node [target,  right of=punkt,  node distance=3.5cm] (pu) {$ (...) $};
        \node [target,  right of=stepm,  node distance=3.5cm] (targetm) {$ \hat{y}_m $};

        \path [line] (step0) -- (target0);
        \path [line] (step1) -- (target1);
        \path [line] (step2) -- (target2);
        \path [line] (step3) -- (target3);

        \path [line, gray, dashed] (punkt) -- (pu);

        \path [line] (stepm) -- (targetm);

        \end{tikzpicture}
    \end{center}
    \caption{Illustration of a cascade of predictors}
    \label{fig:cascade}
\end{figure}

Each of the $m+1$ probability estimators in the product above is a classifier in itself. The feature space of $p(y_k|\mathbf{x}, m, y_1,\ldots,y_{k-1})$ is $\mathcal{X}\times\mathbb{N}\times\mathcal{Y}^{k-1}$. We call such a structure a \emph{cascade} of predictors. The cascade structure unfolds seamlessly from the application of the chain rule (see Figure~\ref{fig:cascade}). In a cascade, one predicts the number of elements in the subset, then the first value of the subset, then the second using the results from the computation of the first, etc.

The name \emph{cascade} comes naturally for such a sequence of predictors. It also refers to the \emph{detector cascade architecture} introduced by \citep{viola01}, which has been subject to some recent attention by \citep{saberian14} for instance.

\subsection{Cascade of NBCs}
Up to this point in this reasoning, no assumption has been made regarding those probability estimators used in the cascade of predictors. Any efficient classification algorithm can be used to predict each level in the cascade. This implies storing in memory $|\mathcal{Y}|+1$ classifiers having increasingly complex feature spaces and predicting values in a class set of size $|\mathcal{Y}|$. Although this might be practically feasible in cases where $|\mathcal{Y}|$ is small enough, it may not scale up to large label sets. Furthermore, the feature spaces of the last predictors in the cascade are complex, requiring powerful learning architectures. Finally, the training of each level in the cascade can take very long. To illustrate this last point, let us consider a training example $(\mathbf{x},\bar{y})$ where $\bar{y}$ has size $m$. Let $k$ be an integer smaller than $m$. The $k$-th level in the cascade estimates the probability that the $k$-th label is a certain $y \in \mathcal{Y}$, given the input features and $k-1$ other labels in the predicted set. During the training phase, this implies that from a single $(\mathbf{x},\bar{y})$, one can derive $m\cdot \binom{m-1}{k-1}$ such training examples ($m$ target labels and $\binom{m-1}{k-1}$ subsets of size $k-1$ taken among the $m-1$ remaining elements of $\bar{y}$). In practice, it implies that from a single training example $(\mathbf{x},\bar{y})$ with a label set of size $m$, the number of training samples presented at the levels of the cascade close to $\frac{m}{2}$ can become very large. To fix ideas, let us use the previous power plants example, with samples having $\bar{y}$ subsets of constant size $m=30$ and let us consider the 15th predictor in the cascade. Then each training example $(\mathbf{x},\bar{y})$ generates $m\cdot \binom{m-1}{k-1} \sim 2\cdot10^9$ examples that need to be processed by the 15th classifier in the cascade during the training phase. Such values can considerably slow down the training of the cascade architecture and become prohibitive in practice.

Now let us discuss how taking NBCs as base classifiers for each level in the cascade induce a dramatic simplification of both training and storage of the multi-label classifier. Let us suppose that each of these estimators is built upon the Naive Bayes assumption. Based on the conclusions of \citet{zhang2004,zhang2005}, although the probability estimates of these classifiers are poor, at each step of the cascade the computed $\argmax$ remains close to optimal.
Eventually, we are left with $|\mathcal{Y}|+1$ NBCs: one for the subset size prediction and one for each level in the cascade. If we start the numbering at zero, predictor zero estimates $\mathbb{P}(M|\mathbf{X})$, then predictor one estimates $p\left(y_1|\mathbf{x},m\right)$, predictor two estimates $p\left(y_{2}|\mathbf{x},m,y_1\right)$ and so on. Let $f_k$ be the selection function of predictor number $k$.

Let us consider the $k$-th predictor in the cascade. Since it is a Naive Bayes classifier, according to Equation \ref{eq:nbc-cont}, its selection function decomposes as

\begin{align}
f_k(\mathbf{x},m,y_1,\ldots,y_{k-1}) =& \argmax_{y_{k}\in\mathcal{Y}} \mathbb{P}(y_k\in Y) \times \nonumber \\
& \mathbb{P}(M=m|y_k\in Y) \times \nonumber \\
& \prod\limits_{i=1}^n p(X_i=x_i|y_k\in Y) \times \nonumber \\
& \prod\limits_{j=1}^{k-1} \mathbb{P}(y_j\in Y|y_k\in Y) \times \nonumber \\
& \frac{1}{p(\mathbf{X}=\mathbf{x},M=m,\{y_1,\ldots y_{k-1}\}\subset Y)}.
\label{eq:final}
\end{align}

Note that finding the $\argmax$ in Equation \ref{eq:final} does not require computing the denominator since it does not depend on $y_k$. We keep this denominator nonetheless for the sake of rigor and because Equation \ref{eq:final} can be used in an extension to our algorithm which is discussed in the perspectives (but beyond the scope of this paper).

Computing the selection function $f_k$ requires evaluating each term of the numerator in Equation \ref{eq:final}. Each term in this numerator is a univariate probability estimator. These estimators are not specific to the $k$-th step in the cascade: take two predictors $f_k$ and $f_{k'}$, both will make use of the same generic probability estimators $\mathbb{P}(y \in Y)$, $p(X_i|y \in Y)$, $\mathbb{P}(M|y\in Y)$ and $\mathbb{P}(y'\in Y|y\in Y)$. The same univariate probability estimators are simply combined in different fashions at the different stages of the cascade.

The cases of $f_1$ and $f_0$ require different computations. Recall that $f_1(\mathbf{x},m)$ is the selection function of the first label. Its computation makes use of the same $\mathbb{P}(y \in Y)$, $p(X_i|y \in Y)$ and $\mathbb{P}(M|y\in Y)$ probability estimators as the rest of the cascade (simply it does not use the $\mathbb{P}(y'\in Y|y\in Y)$ estimator). Finally, $f_0(\mathbf{x})$ selects the most probable subset size associated to $\mathbf{x}$. Since $f_0$ is also a Naive Bayes Classifier, its computation requires the $\mathbb{P}(M)$ and $p(X_i|M)$ univariate probability estimators.

Table \ref{tab:univ_proba} summarizes the 6 univariate distributions that are required for the computation of all levels in the cascade, along with the space complexity of their storage (detailed in section \ref{sec:complex}). Finally, the overall space requirements for the whole cascade is $\mathcal{O}((n\kappa+|\mathcal{Y}|)|\mathcal{Y}|)$.

\begin{table}
    \begin{center}
        \begin{tabular}{ll}
        \toprule
        Distribution & Space complexity \\
        \midrule
        $\mathbb{P}(M)$ & $\mathcal{O}\left(|\mathcal{Y}|\right)$\\
        $p(X_i|M)$ & $\mathcal{O}\left(n\kappa|\mathcal{Y}|\right)$\\
        $\mathbb{P}(y \in Y)$ & $\mathcal{O}\left(|\mathcal{Y}|\right)$\\
        $p(X_i|y \in Y)$ & $\mathcal{O}\left(n \kappa|\mathcal{Y}|\right)$\\
        $\mathbb{P}(M|y\in Y)$ & $\mathcal{O}\left(|\mathcal{Y}|^2\right)$\\
        $\mathbb{P}(y'\in Y|y\in Y)$ & $\mathcal{O}\left(|\mathcal{Y}|^2\right)$\\
        \bottomrule
        \end{tabular}
    \end{center}
    \caption{Univariate distributions in a cascade of NBCs}
    \label{tab:univ_proba}
\end{table}

\section{The \nbx Algorithm}
\label{sec:naibx}

The previous Section set down three concepts:
\begin{itemize}
\item Decomposing the multi-label classification problem into a sequence of single-label predictors, organized in a cascade to retain the dependency relationships between labels.
\item Illustrating why, despite the reduction from multi-label to cascade of single-label predictions, its application can be difficult for large problems.
\item Showing how the introduction of NBCs in a cascade of predictors did not require the estimation of a different classifier at each level of the cascade, but rather made use of 6 generic univariate probability estimators which can be easily computed from available data and which are combined together in different manners by each level of the cascade.
\end{itemize}
In this Section we introduce the \nbx\footnote{\nbx \textipa{(/"neIbEks/)} stands for Naive Bayes Classification for Subset Selection.} algorithm, which computes the distributions introduced in the previous analysis and uses them to make multi-label predictions given new data.

\subsection{Algorithm Overview}
\label{sub:nbx_algo}
\nbx is an online algorithm that combines a training function \texttt{add\_example} and a prediction function \texttt{predict\_y}. It makes use of the data structures presented in Table \ref{tab:data} to store the parameters necessary to the evaluation of the distributions summarized in Table \ref{tab:univ_proba}. Algorithm \ref{algo:naibx-learn} presents the incremental learning process of the \texttt{add\_example} function while Algorithm \ref{algo:naibx-predict} presents the operations performed when a new sample $\mathbf{x}$ requires the prediction of the associated subset of labels.

\begin{table*}
    \begin{center}
        \resizebox{\textwidth}{!}
        {
        \begin{tabular}{lll}
        \toprule
        \textbf{Name}       & \textbf{Description}      & \textbf{Size}\\
        \toprule
        $N$ & total number of examples ``$(\mathbf{x};\bar{y})$'' & $1\times 1$\\
        \midrule
        $N_y$ & array storing the number of examples seen for $y\in Y$ & $1\times|\mathcal{Y}|$\\
        \midrule
        $N_m$ & array storing the number of examples seen for $M=m$ & $1\times(|\mathcal{Y}|+1)$\\
        \midrule
        $\mu_{iy}$ & stores the empirical mean of $\mathbb{P}(X_i=x_i|y\in Y)$ & $n\times|\mathcal{Y}|$\\
        \midrule
        ${M_2}_{iy}$ & table storing the sum of squares of variations to $\mu_{iy}$, used & $n\times|\mathcal{Y}|$\\
        & to compute the standard deviation of $\mathbb{P}(X_i=x_i|y\in Y)$ & \\
        \midrule
        $\mu_{im}$ & stores the empirical mean of $\mathbb{P}(X_i=x_i|M=m)$ & $n\times(|\mathcal{Y}|+1)$\\
        \midrule
        ${M_2}_{im}$ & table storing the sum of squares of variations to $\mu_{im}$, used & $n\times(|\mathcal{Y}|+1)$\\
        & to compute the standard deviation of $\mathbb{P}(X_i=x_i|M=m)$ & \\
        \midrule
        $N_{yy'}$ & table counting the number of occurences of $(y'\in Y|y\in Y)$ & $|\mathcal{Y}|\times|\mathcal{Y}|$\\
        \midrule
        $N_{ym}$ & table counting the number of occurences of $(M=m|y\in Y)$ & $|\mathcal{Y}|\times(|\mathcal{Y}|+1)$\\
        \bottomrule
        \end{tabular}
        }
    \end{center}
\caption{Data structures}
\label{tab:data}
\end{table*}

The \texttt{add\_example} procedure computes the statistics describing the 6 probability distributions required by \nbx for future predictions (presented in Table \ref{tab:univ_proba}). The use of $\mu_{iy}$ and ${M_2}_{iy}$ (resp. $\mu_{im}$ and ${M_2}_{im}$) is recommended by \citep{knuth1997,chan83} for the online calculation of mean and variance of a population.

The \texttt{predict\_y} function receives a new observation $\mathbf{x}_{new}$ as an input and predicts a vector of labels $\mathbf{y}_{pred}$ in a two-step process. In the first one, it estimates the size of the target vector via the \texttt{predict\_m} function. In the second step it proceeds by estimating the elements of the vector through the cascade of predictors. At each iteration the function \texttt{predict\_yk} is called and fed as an input the size $\hat{m}$ and the labels estimated so far.
Note that \nbx was thought as the natural extension of Naive Bayes Classifiers to the multi-label case. If one trains \nbx on a data set with targets $\mathbf{y}_{obs}$ of size $m = 1$ with values from a target set of size $|\mathcal{Y}| = 2$, that is
    $$f_{classifier} : \mathcal{X} \longrightarrow \mathcal{Y} = \{c_1, c_2\},
    $$
    then \nbx will act as a traditional binary Naive Bayes Classifier. Furthermore, allowing $|\mathcal{Y}| > 2$ will return a multi-class classifier.

\begin{algorithm}
    \DontPrintSemicolon
    \underline{add\_example($\mathbf{x},\mathbf{y}$)} \;
    $m=$ length of $\mathbf{y}$\;
    $N \leftarrow N+1$\;
    $N_m \leftarrow N_m+1$\;
    \For{$i=1$ to $n$}{
      $\delta = x_i - \mu_{im}$\;
      $\mu_{im} \leftarrow \mu_{im} + \frac{\delta}{N_m}$\;
      ${M_2}_{im} \leftarrow {M_2}_{im} + \delta \times \left(x_i - \mu_{im}\right)$\;
    }
    \For{$j=1$ to $m$}{
      $y = \mathbf{y}[j]$ \tcc*{target value}


      $N_y \leftarrow N_y+1$\;
      \For{$i=1$ to $n$}{
        $\delta = x_i - \mu_{iy}$\;
        $\mu_{iy} \leftarrow \mu_{iy} + \frac{\delta}{N_y}$\;
        ${M_2}_{iy} \leftarrow {M_2}_{iy} + \delta \times \left(x_i - \mu_{iy}\right)$\;
      }
      \For{$k=1$ to length of $\mathbf{y}$, $k\neq j$}{
        $y' = \mathbf{y}[k]$ \tcc*{feature value}
        $N_{yy'} \leftarrow N_{yy'}+1$\;
      }
    }
    \caption{\nbx---Learning Step}
    \label{algo:naibx-learn}
\end{algorithm}

\begin{algorithm}
    \DontPrintSemicolon

    \underline{predict\_y($\mathbf{x}_{new}$)} \;
    $m$ = predict\_m$(\mathbf{x}_{new})$\;
    $\mathbf{y}_{pred}=\emptyset$\;

    \For{$k=1$ to $m$}{
      $\mathbf{y}_{pred} \leftarrow \mathbf{y}_{pred} \ \cup $ predict\_yk($\mathbf{x}_{new}, m, \mathbf{y}_{pred}$)\;
    }
    \Return{$\mathbf{y}_{pred}$}\;
    \;

    \underline{predict\_m($\mathbf{x}_{new}$)} \;
    \For{$m=0$ to $|\mathcal{Y}|$}{

      $score_m=\frac{N_m}{N}$\;
      \For{$i=1$ to $n$}{
        $\sigma_{im}^2 = \frac{{M_2}_{im}}{N_m-1}$\;
        $d_{im} = \frac{1}{\sqrt{2\pi\sigma_{im}^2}}\exp\left(\frac{-(\mu_{im}-x_i)^2}{2\sigma_{im}^2}\right)$\;
        $score_m = score_m\times d_{im}$\;
      }
    }
    \Return{$m$ with the best $score_m$}\;
    \;

    \underline{predict\_yk($\mathbf{x}_{new}$, m, $\mathbf{y}_f$)} \;
    $k$ = 1 + length of $\mathbf{y}_f$\;
    \For{$y \in \mathcal{Y} \setminus \{\mathbf{y}_f\}$}{
      $p_y = \frac{N_y}{N}$\;

      $score_y = p_y$\;
      \For{$i=1$ to $n$}{
        $\sigma_{iy}^2 = \frac{{M_2}_{iy}}{N_y-1}$\;
        $d_{iy} = \frac{1}{\sqrt{2\pi\sigma_{iy}^2}}
            \exp\left(\frac{-(\mu_{iy}-x_i)^2}{2\sigma_{iy}^2}\right)$\;
        $score_y = score_y\times d_{iy}$\;
      }
      $p_{my} = \frac{N_{ym}}{N_y}$\;
      $score_y = score_y\times p_{my}$\;
      \For{$j=1$ to $k-1$}{
        $y'={y_f}_j$\;
        $p_{jy} = \frac{N_{yy'}}{N_y}$\;
        $score_y = score_y\times p_{jy}$\;
      }
    }
    Normalize scores \;
    \Return{$y$ with the best $score_y$}\;

    \caption{\nbx --- Prediction Step}
    \label{algo:naibx-predict}
\end{algorithm}

The algorithm presented in Algorithms \ref{algo:naibx-learn} and \ref{algo:naibx-predict} assumes a Gaussian distribution for all continuous variables. However, \nbx can be straightforwardly extended to any other type of distribution.

\subsection{Complexity Analysis}
\label{sec:complex}

Storing the cascade of predictors during the training phase boils down to storing the parameters of the six probability distributions presented in Tables \ref{tab:univ_proba} and \ref{tab:data}. The space complexity of storing these parameters are recalled in the above tables. Thus, as stated before, the space requirements of the whole cascade of classifiers is in $\mathcal{O}\left(|\mathcal{Y}|\left(\kappa n + |\mathcal{Y}|\right)\right)$.

As a matter of illustration, with $|\mathcal{Y}|=150$, $m=30$ (as in the power plants example), $\kappa=2$ and $n = 10000$, this upper bound is in the order of $10^6$: storing $10^6$ values seems much more practical than the estimated $10^{63}$ of the initial example (see Section \ref{mot_ex}).

It is also relevant to note that if one specializes the previous approach to the prediction of fixed-length subsets of size $m$ (as in the power plants example), then the analysis above still holds with the simplification that there is no need for predictor number zero.

The time complexity of the training and prediction phases unfolds straightforwardly from the presentation in Algorithms \ref{algo:naibx-learn} and \ref{algo:naibx-predict}. These remarks are summarized in Table \ref{tab:nbx_compl}.

\begin{table}
    \begin{center}
        \begin{tabular}{ccc}
        \toprule
         & Training & Prediction\\
        \midrule
        Time
            & $\mathcal{O}\left(|\mathcal{Y}|\left(\kappa n+|\mathcal{Y}|\right)\right)$
                & $\mathcal{O}\left(|\mathcal{Y}|^{2} \left(n + |\mathcal{Y}| \right)\right)$\\

        Space
            & $\mathcal{O}\left(|\mathcal{Y}|\left(|\mathcal{Y}| + n\right)\right)$
                & $\mathcal{O}(|\mathcal{Y}|)$ \\
        \bottomrule
        \end{tabular}
    \end{center}
\caption{\nbx---Time and Space Complexity}
\label{tab:nbx_compl}
\end{table}

\subsection{Laplacian Smoothing}
A common practice in Naive Bayes prediction is to use additive (Laplacian) smoothing in the estimation of probabilities related to discrete variables, in order to compensate for unseen examples and to avoid having zero estimates for $\mathbb{P}(X|Y)$ and $\mathbb{P}(Y)$ that pull the estimated probability of a class down to zero.
In the general case, for a discrete variable $Y$ which can take $K$ possible values and for which $Y=y$ has been observed $N_y$ times after $N$ observations, estimating $\mathbb{P}(Y=y)$ using Laplacian smoothing can be written
\begin{equation}
\mathbb{P}(Y=y) = \frac{N_y+1}{N+K}.
\label{eq:laplace}
\end{equation}

This sample correction corresponds to supposing that each possible value of $Y$ has been seen exactly once before any other observation, which implies that $\mathbb{P}(Y)$ is non-zero. Asymptotically, this does not affect the overall behaviour of the probability estimator.
In our analyses we applied it when estimating the probabilities $\mathbb{P}(M=m)$, $\mathbb{P}(Y=y)$, $\mathbb{P}(M=m|Y=y)$ and $\mathbb{P}(Y'=y'|Y=y)$.
In our case, attention must be paid in the case of $\mathbb{P}(Y'=y'|Y=y)$.
One knows that $\mathbb{P}(Y'=y'|Y=y)=0$ if $y=y'$. This violates the assumption behind Laplacian smoothing that every possible value of $Y'$ has been observed exactly once before sample collection started. Consequently, the estimation of $\mathbb{P}(Y'=y'|Y=y)$ should follow Equation \ref{eq:laplaceY}. Note in particular that we use $K = |\mathcal{Y}|-1$ since there is one forbidden value for $Y'$.

\begin{equation}
    \mathbb{P}(Y'=y'|Y=y) = \left\{\begin{array}{ll}
    0 & \textrm{if } y=y'\\
    \frac{N_{yy'}+1}{N_y+|\mathcal{Y}|-1} & \textrm{otherwise}
    \end{array}\right.
    \label{eq:laplaceY}
\end{equation}

\subsection{Bag-of-Words Text Representation}
\label{sec:bow}
Throughout the paper, the features are assumed continuous, as this was the framework within which \nbx was thought of and developed. Nonetheless, as in the MLC literature text analysis is one of the most relevant applications, we provide some examples of application to such data sets. In our implementation for \nbx, the Naive Bayes classifiers are built following a standard approach, as seen for example in \citet{retrieval}. Each entry in the input data is a Bag-of-Words (\textit{BoW}) representation of a document, that is, each keyword appearing in the document is associated to the number of times it appears in the text. When indicated as ``binary'', for each observation we only have information on whether or not a word was included. In this latter case a Bernoulli model is used, while for the former we opted for a multinomial model.
We kept the estimation of probabilities to the prediction step and limited the training phase to updating the tables of frequencies.

\section{Experimental Results}
\label{sec:experiments}

In this Section we introduce applications of \nbx  to real data sets, the methodologies followed and observations on the results obtained. We perform a comparison of predictive performance (computational time and prediction quality) against some popular models implemented in MEKA \citep{MEKA}, adopting the default parameters.

\subsection{Data Sets}

\begin{table}
    \centering
    \begin{tabular}{l r r r r}
        \multicolumn{5}{c}{Continuous Features}       \\
        \toprule
        \textbf{data set}    & $N$    & $dim(\mathcal{X})$  & $Labels$  & $LCard$ \\
        \midrule
                \textit{CAL500}      &  502   & 68    & 174  &    {26.043}        \\
                \textit{Emotions}    &  593   &  72   & 6    &    1.869         \\
                \textit{Mediamill}   & 43902  & 120   & 101  &    4.376         \\
                \textit{Music}       & 592    & 71    & 6    &    1.870         \\
                \textit{NUS-WIDE}    & {269648} & 128   & 81   &    1.873         \\
                \textit{Scene}       & 2407   & 294   & 6    &    1.074         \\
                \textit{Yeast}       & 2417   & 103   & 14   &    4.237         \\
         \bottomrule
    \end{tabular}
    \caption{Description---Continuous data sets}
    \label{tab:cont}
\end{table}

\begin{table}
    \centering
    \begin{tabular}{l r r r r}

        \multicolumn{5}{c}{Bag-of-Words Features}       \\
        \toprule
        \textbf{data set}    & $N$    & $dim(\mathcal{X})$  & $Labels$  & $LCard$ \\
        \midrule
                {Bibtex}      & 7395   & 1836 & 159 & 2.402 \\
                {Enron}       & 1702   & 1001 & 53  & 3.378 \\
                {LLog}        & 1460   & 1004 & 75  & 1.180 \\
                {Slashdot}    & 3782   & 1079 & 22  & 1.181 \\                
                \midrule
                {{Yahoo}}       &    & &  &             \\
                1)  Arts          & 7484  & 23146 & 26 & 1.654 \\
                2)  Business      & 11214 & 21924 & 30 & 1.599 \\
                3)  Computers     & 12444 & 34096 & 33 & 1.507 \\
                4)  Education     & 12030 & 27534 & 33 & 1.463 \\
                5)  Entertainment & 12730 & 32001 & 21 & 1.414 \\
                6)  Health        & 9205  & 30605 & 32 & 1.644 \\
                7)  Recreation    & 12828 & 30324 & 22 & 1.429 \\
                8)  Reference     & 8027  & 39679 & 33 & 1.174 \\
                9)  Science       & 6428  & 37187 & 40 & 1.450 \\
                10) Social        & 12111 & 52350 & 39 & 1.279 \\
                11) Society       & 14512 & 31802 & 27 & 1.670 \\
         \bottomrule
    \end{tabular}
    \caption{Description---Text data sets}
    \label{tab:bowdata}
\end{table}

As stated previously (Section \ref{mot_ex}), our work stems from a practical problem encountered in optimization in engineering and our direct interest is on continuous features. Nonetheless, as \nbx is easily extensible to handle discrete features in the form of Bag-of-Words (Section \ref{sec:bow}) we provide some application examples.
The computational experiments were conducted on data taken from a set of standard MLC data sets\footnote{See \url{http://mulan.sourceforge.net/data sets-mlc.html} and \url{http://meka.sourceforge.net/}.} commonly adopted in the literature.
We worked on all the examples with continuous features and those with a Bag-of-Words encoding. In Tables \ref{tab:cont} and \ref{tab:bowdata} one can find the list of the input data.

The \textit{Yahoo} data set was split into 11 subsets at the source and, according to practice in the literature, we provide analysis for each of them individually.

\subsection{Evaluation Metrics}

As ``\textit{not all data sets are equally multi-label}'' \citep{tsoumakas07}, one needs to quantify multi-labelness. A common metric of choice is the average dimension of the target vectors across the data set, known as \textit{label cardinality}

$$ LCard = \frac{1}{N} \sum_{i=1}^{N} \left|\textbf{y}_{i}^{obs}\right|.  $$

While in binary classification one can directly look at the number of times the prediction was equal to the observed class and thus get an idea of the quality of the model, for MLC the issue is not as straighforward.

One could adopt the \textit{zero-one} loss metric, common in the classification literature \citep{tsoumakas07}

$$ L_{01} = 1 - \frac{1}{N} \sum_{i=1}^{N}\mathbb{I}(\textbf{y}^i = \textbf{\^{y}}^i),$$

or alternatively the \textit{zero-one} score
$$ Z_{S} =     \frac{1}{N} \sum_{i=1}^{N}\mathbb{I}(\textbf{y}^i = \textbf{\^{y}}^i),$$

where $\textbf{y}^i$ and $\textbf{\^{y}}^i$ are respectively the observed  and predicted target vectors for the $i$-th observation, and $\mathbb{I}(\textbf{y}^i = \textbf{\^{y}}^i)$ takes value one if the two vectors are exactly the same and zero otherwise.

In the case of targets of dimension one ($m = 1$) this corresponds to the error rate \citep[see][chap.~7]{hastie2001}. In MLC, this returns the \textit{proportion} of samples correctly predicted. For very large target vectors $\mathbf{y_i}$, the zero-one Loss becomes less meaningful as a single mistake will invalidate an otherwise good prediction while, for example, one could see as a success the correct prediction of 23 out of 25 labels. We can introduce the notion of \textit{partially correct} prediction. Intuitively, we want to measure the average number of operations it would take to turn the predicted vector $\mathbf{y}$ into the correct one. If the prediction is perfect, then this will return zero. This is captured by the Hamming Loss metric, defined as

	$$ HL = 1 - \frac{1}{N} \sum_{i=1}^{N}
						\left(\frac{1}{L} \sum_{k=1}^{L}
							\mathbb{I}\left(y_k^i = \widehat{y}_k^i\right)\right)  \in [0,1],    $$

or alternatively as the Hamming score
    $$ H_S = \frac{1}{N} \sum_{i=1}^{N}
						\left(\frac{1}{L} \sum_{k=1}^{L}
							\mathbb{I}\left(y_k^i = \widehat{y}_k^i\right)\right)  \in [0,1],    $$

    where $L$ is the number of available labels.

In the literature, other commonly found measures are Accuracy, Precision and Recall \citep{tsoumakas07} given respectively by

\begin{align*}
\mathit{Acc} =
	\frac{1}{N}
		\sum_{i=1}^{N}
		\frac{\left | \textbf{y}^i \cap \widehat{\textbf{y}}^i \right |}
    {\left | \textbf{y}^i \cup \widehat{\textbf{y}}^i \right |},           &  &\mathit{Pre} =
	\frac{1}{N}
		\sum_{i=1}^{N}
		\frac{\left | \textbf{y}^i \cap \widehat{\textbf{y}}^i \right |}
		 {\left | \widehat{\textbf{y}}^i \right |},              &  &\mathit{Rec} =
	\frac{1}{N}
	\sum_{i=1}^{N}
	\frac{\left | \textbf{y}^i \cap \widehat{\textbf{y}}^i \right |}
	{\left | \textbf{y}^i \right |}.
\end{align*}

%
%

None of these metrics can be considered flawless. In the case of data sets with low cardinality and big $|\mathcal{Y}|$ for  example, the great majority of labels are not included in the predicted vector and for many labels we have $y_i^j = 0$. If the predicted labels are all wrong we can still get small values of $H_S$. On the other hand, if one works with \textit{Recall}, one will have an advantage at predicting as many labels as possible. This will give us high values of recall at the expense of many labels predicted when not needed, that is, many false positives will be predicted. In general, measuring the performance in MLC can be a problem in itself, as some models can perform better than others given a specific metric.

\subsection{Methods}
In our model, we can distinguish two main features: the choice of a cascade of predictors to perfom multi-label classification and the choice to adopt Naive Bayes Classifier as label-estimation tool.
Among the algorithms available for MLC, the chains of classifiers (CC) and their \textit{ensemble} (ECC) extension introduced by \citet{read2009} seem to be one of the most efficient variations on BR, while RAkEL \citep{Tsoumakas2011} is among the most interesting variations on LP. We therefore compare our model to these, making prediction via CC and ECC with SVM and NBC as base predictors (referred to as \svm \& \esvm and \nbc \& \enbc) and RAkEL with SVM as base predictor (default parameters: $k = 3$, $m = 10$).
As exposed in Section \ref{mot_ex}, we are interested in methods granting running times in the order of seconds or minutes. Nonetheless, for the sake of comparing predictive performances, we allowed running times (including handling data, training and testing) of up to 12 hours with 4 GB of memory reserved to the task\footnote{Experiments were run on a 64bit Intel i7-4800MQ @ 2.70GHz architecture, with 16 GB of RAM, operating under a GNU/Linux Fedora 24 OS.}. For methods that failed to deliver a result, their corresponding line in the tables is left blank. In the case of the \textit{Yahoo} data sets, neither $ECC_{SVM/NBC}$ nor \rk could produce any result and we thus we restricted the comparison to \svm and \nbc. We adopted the default parameters for all the algorithms applied in the analyses.

When possible, the models were cross-validated with a 10-fold method \citep[see][chap.~7]{hastie2001} and the estimated losses were obtained  as $     \bar{\ell} = \frac{1}{10}\sum_{k=1}^{10}{\ell}_k. $ For bigger data sets (including \textit{Yahoo}), as most of the algorithms would produce no output within our time and space limits, we split the data into training and testing partitions, reserving 66\% of the data to training (indicated in the tables as ``CV 66\%'').
Computing times are included to assess the impact of the algorithms' complexity on performances and are indicated in the tables as $T_{train}$ and $T_{test}$, respectively the number of seconds spent on the training and prediction steps.

For the prediction step, \nbx is composed of two distinct phases (Section \ref{sub:nbx_algo}), the prediction of the size of the target and the following prediction of the labels via the cascade scheme. The prediction of the target size $m$ is a peculiar feature of \nbx and we think that it deserves some attention. We applied our algorithm skipping the first step of size prediction by using as an input the true, observed, size $m$ and reported the results as $NBX_{True M}$, (details in Section \ref{sec:truem}).

\subsection{Results for Continuous Features Data Sets}
\label{sec:cont_results}

\begin{table}
\centering

\resizebox{\textwidth}{!} {
\begin{tabular}{@{}lrrrrrrrrrr@{}}
\toprule
\multicolumn{1}{c}{{Data}}
    & \multicolumn{1}{c}{{ALGO}}
    & \multicolumn{1}{c}{\textbf{$LCard$}}
    & \multicolumn{1}{c}{\textbf{$\widehat{LCard}$}}
    & \multicolumn{1}{c}{\textbf{$H_s$}}
    & \multicolumn{1}{c}{\textbf{$Z_s$}}
    & \multicolumn{1}{c}{\textbf{$Acc$}}
    & \multicolumn{1}{c}{\textbf{$Pre$}}
    & \multicolumn{1}{c}{\textbf{$Rec$}}
    & \multicolumn{1}{c}{\textbf{$T_{train}[s]$}}
    & \multicolumn{1}{c}{\textbf{$T_{pred}[s]$}} \\
\midrule
Yeast     & \svm  & 4.237  & 4.245   & 0.785 & 0.201 & 0.529 & 0.652   & 0.660   & 3.517     & 0.011   \\
          & \esvm &        & 3.949   & \textbf{0.799} & \textbf{0.203} & 0.536 & \textbf{0.680}   & 0.633   & 30.826    & 0.138   \\
          & \nbc  &        & 6.165   & 0.706 & 0.096 & 0.476 & 0.545   & \textbf{0.744}   & 0.312     & 0.149   \\
          & \enbc &        & 5.122   & 0.701 & 0.105 & 0.428 & 0.505   & 0.610   & 2.056     & 1.313   \\
          & \rk   &        & \textbf{4.244}   & 0.796 & 0.180 & \textbf{0.541} & 0.663   & 0.664   & 12.848    & 0.020   \\
          & \nbx  &        & 4.551   & 0.705 & 0.115 & 0.405 & 0.541   & 0.555   & 0.041     & 0.629   \\
          \mline
          & \tm   &        &    -     & 0.757 & 0.218 & 0.481 & 0.579   & 0.579   & 0.043     & 0.502   \\
          \midrule
Scene     & \svm  & 1.074  & \textbf{1.091}   & 0.892 & 0.634 & 0.688 & 0.711   & 0.720   & 3.288     & 0.007   \\
          & \esvm &        & 1.013   & \textbf{0.906} & \textbf{0.642} & \textbf{0.694} & \textbf{0.752}   & 0.710   & 20.154    & 0.078   \\
          & \nbc  &        & 2.174   & 0.762 & 0.172 & 0.454 & 0.461   & \textbf{0.854}   & 0.383     & 0.172   \\
          & \enbc &        & 2.160   & 0.765 & 0.178 & 0.460 & 0.422   & 0.848   & 2.307     & 1.483   \\
          & \rk   &        & 1.369   & 0.885 & 0.538 & 0.684 & 0.640   & 0.816   & 23.808    & 0.028   \\
          & \nbx  &        & 1.408   & 0.866 & 0.453 & 0.623 & 0.631   & 0.784   & 0.050     & 0.277   \\
          \mline
          & \tm   &        &    -     & 0.903 & 0.712 & 0.721 & 0.725   & 0.725   & 0.050     & 0.166   \\
          \midrule
Emotions  & \svm  & 1.868  & 2.066   & 0.780 & 0.283 & 0.550 & 0.633   & 0.702   & 0.131     & 0.001   \\
          & \esvm &        & \textbf{1.934}   & \textbf{0.806} & \textbf{0.320} & 0.578 & \textbf{0.682}   & 0.706   & 0.695     & 0.006   \\
          & \nbc  &        & 2.551   & 0.739 & 0.211 & 0.521 & 0.565   & 0.766   & 0.017     & 0.009   \\
          & \enbc &        & 2.447   & 0.755 & 0.233 & 0.538 & 0.581   & 0.761   & 0.125     & 0.102   \\
          & \rk   &        & 2.245   & 0.795 & 0.317 & \textbf{0.592} & 0.642   & \textbf{0.771}   & 0.759     & 0.003   \\
          & \nbx  &        & 1.947   & 0.771 & 0.275 & 0.528 & 0.639   & 0.648   & 0.000     & 0.017   \\
          \mline
          & \tm   &        &    -     & 0.814 & 0.526 & 0.623 & 0.669   & 0.669   & 0.000     & 0.013   \\
          \midrule
CAL500    & \svm  & 26.044 & 23.906  & 0.823 & 0.000 & 0.244 & 0.412   & 0.375   & 5.866     & 0.227   \\
          & \esvm &        & 13.572  & 0.855 & 0.000 & 0.228 & 0.531   & 0.277   & 29.265    & 3.427   \\
          & \nbc  &        & 78.195  & 0.571 & 0.000 & 0.180 & 0.209   & \textbf{0.562}   & 2.398     & 0.521   \\
          & \enbc &        & 56.779  & 0.690 & 0.000 & 0.222 & 0.254   & 0.554   & 3.484     & 6.567   \\
          & \rk   &        & 10.020  & \textbf{0.863} & 0.000 & 0.209 & \textbf{0.610}   & 0.235   & 24.480    & 0.142   \\
          & \nbx  &        & \textbf{25.719}  & 0.817 & \textbf{0.002} & \textbf{0.249} & 0.379   & 0.372   & 0.031     & 6.235   \\
          \mline
          & \tm   &        &     -    & 0.833 & 0.012 & 0.299 & 0.423   & 0.423   & 0.031     & 6.285   \\
          \midrule
Mediamill & \svm  & 4.406  & \textbf{2.915}   & \textbf{0.963} & \textbf{0.057} & \textbf{0.368} & 0.605   & 0.455   & 13275.522 & 20.986  \\
(CV 66\%)   & \esvm &        &         &       &       &       &         &         &           &         \\
          & \nbc  &        & 41.738  & 0.604 & 0.000 & 0.073 & 0.074   & \textbf{0.697}   & 67.211    & 86.443  \\
          & \enbc &        & 24.330  & 0.769 & 0.000 & 0.110 & 0.112   & 0.616   & 376.396   & 924.879 \\
          & \rk   &        & 0.285   & 0.957 & 0.055 & 0.092 & \textbf{0.658}   & 0.043   & 17486.113 & 3.392   \\
          & \nbx  &        & 8.033   & 0.909 & 0.014 & 0.134 & 0.174   & 0.339   & 0.086     & 44.099  \\
          \mline
          & \tm   &        &     -    & 0.932 & 0.048 & 0.110 & 0.168   & 0.168   & 0.086     & 23.631  \\
          \midrule
Music     & \svm  & 1.870  & 2.059   & 0.786 & 0.304 & 0.564 & 0.650   & 0.715   & 0.083     & <0.001   \\
          & \esvm &        & \textbf{1.895}   & \textbf{0.807} & \textbf{0.326} & 0.577 & \textbf{0.687}   & 0.696   & 0.689     & 0.006   \\
          & \nbc  &        & 2.544   & 0.746 & 0.215 & 0.527 & 0.570   & \textbf{0.774}   & 0.027     & 0.011   \\
          & \enbc &        & 2.448   & 0.754 & 0.240 & 0.538 & 0.580   & 0.760  & 0.122  & 0.100   \\
          & \rk   &        & 2.230   & 0.794 & 0.314 & \textbf{0.592} & 0.642   & 0.766  & 0.948     & 0.003   \\
          & \nbx  &        & 1.926   & 0.771 & 0.284 & 0.530 & 0.643   & 0.643  & <0.001 & 0.016   \\
          \mline
          & \tm   &        &    -     & 0.815 & 0.526 & 0.624 & 0.671   & 0.671  & <0.001 & 0.013   \\
          \midrule
NUS-WIDE  & \svm  & 1.863  & 0.484   & \textbf{0.979} & \textbf{0.257} & 0.124 & 0.192   & 0.131   & 41891.741 & 53.344  \\
(CV 66\%) & \esvm &        &         &       &       &       &         &        &           &         \\
          & \nbc  &        & 4.988   & 0.937 & 0.106 & 0.142 & 0.167 & \textbf{0.329} & 336.166   & 245.376 \\
          & \enbc &        &         &       &       &       &         &        &           &         \\
          & \rk   &        &         &       &       &       &         &        &           &         \\
          & \nbx  &        & \textbf{1.627}   & 0.968 & 0.186 & \textbf{0.164} & \textbf{0.223}   & 0.201   & 0.249     & 67.302  \\
          \mline
          & \tm   &        &    -     &  0.970 & 0.356 & 0.223 & 0.263   & 0.264   & 0.249     & 67.389  \\
          
          \midrule
          \multicolumn{2}{r}{$Average\left(\Delta_{Loss}\right)$} & & & -0.045  & -0.068  & -0.077 & -0.147 & -0.170 & &\\
          \multicolumn{2}{c}{{ (\nbx \textit{vs} best opponent)}} & &  & & & & &  & &  \\          
          \bottomrule
\end{tabular} }
\caption{Metrics---Continuous Features}
\label{tab:res_cont}
\end{table}

Albeit with some exceptions, it appears that for the case of continuous features more refined approaches such as \esvm and \rk yield better overall performances as shown in Table \ref{tab:res_cont}. They do so, however, at the expense of increased computational times and storage complexity. While for small data sets this can be neglected, for bigger ones computation costs can become prohibitively expensive. This appears evident in the case of the NUS-WIDE data set: \esvm and \rk could not produce any result after 12 hours of computations and \svm did it after about 11 hours for training and one minute for prediction. \nbx, on the other hand, took 0.249 seconds for training and about a minute for prediction. Even though execution times cannot be compared directly because of the different implementations involved, these relative differences are a good indicator of the limit of these more accurate (on average) tools and the performance-cost trade-off involved. \nbx can scale up to bigger data sets rather seamlessly while retaining the possibility of being adopted as an online algorithm, a property that, for example, SVM-based methods cannot offer.

For high-dimensional data sets, the overhead costs associated to our prototype implementation of \nbx are overcome by the costs due to the complexity of the algorithms. For simpler data sets such as \textit{emotions} and \textit{music} ($|\mathcal{Y}|=6$), \nbx prediction was ten times slower than $CC$ methods. For bigger data set such as \textit{Mediamill} and \textit{NUS-WIDE} (respectively $|\mathcal{Y}|=101$ and $|\mathcal{Y}|=81$) this difference disappears. With \textit{CAL500}, the most complex data set with its 174 labels available, this fact is once again evident: despite being the smallest (502 observations), $ECC$ methods present the same order of prediction times as \nbx while being considerably slower during training. \nbx can be highly competitive for data sets of high cardinality as in this case. Comparing it with \rk and \esvm, we see how severely the complexity of these approaches is affecting the process with respect to other data sets and how \nbx managed to better approximate the average target size and the accuracy.

When reading results on the \textit{recall} criterion, one can see how an algorithm that overestimates $m$ will consistently outperform the others in terms of this metric, as for example observed for \nbc. On the same line, one can observe a direct relation between the predicted cardinality and \textit{precision} values: on average, the lower the cardinality, the higher the score. Table \ref{tab:deltas} presents the average difference between \nbx and the best non-\nbx classifier on each dataset. The previous comments mitigate the seemingly bad results on \textit{precision} and \textit{recall}. The good overall performance of \nbx is very close on average to its best competitor for $H_s$, $Z_s$ and $Acc$, when it is not the best itself (\textit{CAL500} and \textit{NUS-WIDE}).

\subsection{Results for \textit{BoW} Data Sets}
\begin{table}
\centering
\resizebox{\textwidth}{!}
{
\begin{tabular}{@{}lrrrrrrrrrr@{}}
\toprule
\multicolumn{1}{c}{{Data}}
    & \multicolumn{1}{c}{{ALGO}}
        & \multicolumn{1}{c}{\textbf{$LCard$}}
        & \multicolumn{1}{c}{\textbf{$\widehat{LCard}$}}
        & \multicolumn{1}{c}{\textbf{$H_s$}}
        & \multicolumn{1}{c}{\textbf{$Z_s$}}
        & \multicolumn{1}{c}{\textbf{$Acc$}}
        & \multicolumn{1}{c}{\textbf{$Pre$}}
        & \multicolumn{1}{c}{\textbf{$Rec$}}
        & \multicolumn{1}{c}{\textbf{$T_{train} [s]$}}
        & \multicolumn{1}{c}{\textbf{$T_{pred} [s]$}} \\
      \midrule

Bibtex    & \svm  & 2.402 & 1.724  & \textbf{0.985} & \textbf{0.161} & 0.328 & \textbf{0.488} & 0.362 & 309.260    & 30.672  \\
(CV 66\%) & \esvm &       & \textbf{2.465}  & 0.982 & 0.148 & \textbf{0.348} & 0.426 & 0.451 & 1328.649   & 304.649 \\
          & \nbc  &       & 18.830 & 0.882 & 0.074 & 0.190 & 0.084 & \textbf{0.680} & 70.391     & 166.432 \\
          & \enbc &       &        &       &       &       &       &       &            &         \\
          & \rk   &       & 0.274  & 0.984 & 0.022 & 0.050 & 0.369 & 0.043 & 76.862     & 10.585  \\
          & \nbx  &       & 1.280  & 0.984 & 0.147 & 0.278 & 0.461 & 0.301 & 0.244      & 53.126  \\
          \mline
          & \tm   &       & -      & 0.983 & 0.212 & 0.346 & 0.406 & 0.406 & 0.249      & 86.295  \\
          \midrule
Enron     & \svm  & 3.378 & 3.220  & 0.939 & 0.132 & 0.409 & 0.521 & 0.497 & 44.548     & 0.328   \\
          & \esvm &       & 2.958  & \textbf{0.947} & \textbf{0.143} & \textbf{0.448} & \textbf{0.600} & 0.525 & 156.391    & 3.401   \\
          & \nbc  &       & 11.627 & 0.796 & 0.004 & 0.228 & 0.181 & \textbf{0.622 }& 20.205     & 4.710   \\
          & \enbc &       & 10.425 & 0.816 & 0.004 & 0.243 & 0.195 & 0.601 & 117.616    & 47.140  \\
          & \rk   &       & 3.052  & 0.938 & 0.068 & 0.354 & 0.512 & 0.463 & 234.868    & 1.183   \\
          & \nbx  &       & \textbf{3.269}  & 0.923 & 0.016 & 0.267 & 0.397 & 0.431 & 0.897      & 18.638  \\
          \mline
          & \tm   &       & -      & 0.928 & 0.134 & 0.357 & 0.460 & 0.460 & 0.882      & 17.984  \\
          \midrule
LLog      & \svm  & 1.180 & 0.692  & 0.981 & 0.237 & 0.288 & 0.321 & 0.188 & 30.969     & 0.695   \\
          & \esvm &       & 0.356  & \textbf{0.984} & 0.234 & 0.265 & \textbf{0.473} & 0.143 & 135.000    & 5.893   \\
          & \nbc  &       & 25.403 & 0.672 & 0.142 & 0.180 & 0.039 & \textbf{0.850} & 25.726     & 5.612   \\
          & \enbc &       &        &       &       &       &       &       &            &         \\
          & \rk   &       & 0.679  & 0.981 & 0.221 & 0.268 & 0.293 & 0.168 & 140.102    & 1.390   \\
          & \nbx  &       & \textbf{0.859}  & \textbf{0.984} & \textbf{0.420} & \textbf{0.341} & 0.422 & 0.341 & 0.565      & 7.382   \\
          \mline
          & \tm   &       & -      & 0.983 & 0.479 & 0.375 & 0.392 & 0.392 & 0.611      & 7.776   \\
          \midrule
SlashDot  & \svm  & 1.181 & \textbf{1.139}  & 0.951 & 0.413 & \textbf{0.499} & 0.541 & 0.522 & 28.972     & 0.169   \\
          & \esvm &       & 0.941  & 0.957 & 0.411 & 0.489 & \textbf{0.629} & 0.502 & 116.427    & 2.351   \\
          & \nbc  &       & 1.576  & 0.936 & 0.271 & 0.422 & 0.426 & 0.568 & 13.783     & 4.979   \\
          & \enbc &       & 1.494  & 0.940 & 0.281 & 0.431 & 0.451 & 0.570 & 82.915     & 48.31   \\
          & \rk   &       & 0.705  & 0.946 & 0.191 & 0.257 & 0.490 & 0.293 & 30.051     & 0.059   \\
          & \nbx  &       & 0.905  & \textbf{0.961} & \textbf{0.424} & 0.497 & 0.535 & \textbf{0.533} & 0.290      & 4.445   \\
          \mline
          & \tm   &       & -      & 0.970 & 0.680 & 0.702 & 0.713 & 0.713 & 0.294      & 5.389  \\
          
          \midrule
          \multicolumn{2}{r}{$Average\left(\Delta_{Loss}\right)$} & & & -0.005 & 0.013 & -0.050 & -0.094 & -0.279 & &\\
          \multicolumn{2}{c}{{ (\nbx \textit{vs} best opponent)}} & &  & & & & &  & &  \\          
          \bottomrule

\end{tabular}}
\caption{Metrics---Binary {BoW} Features (text data)}
\label{tab:bin_res_bow}
\end{table}

\begin{table}[]
\centering
\resizebox{\textwidth}{!}
{
\begin{tabular}{@{}lrrrrrrrrrr@{}}
\toprule
    \multicolumn{1}{c}{{Data}}
    & \multicolumn{1}{c}{{ALGO}}
    & \multicolumn{1}{c}{\textbf{$LCard$}}
    & \multicolumn{1}{c}{\textbf{$\widehat{LCard}$}}
    & \multicolumn{1}{c}{\textbf{$H_s$}}
    & \multicolumn{1}{c}{\textbf{$Z_s$}}
    & \multicolumn{1}{c}{\textbf{$Acc$}}
    & \multicolumn{1}{c}{\textbf{$Pre$}}
    & \multicolumn{1}{c}{\textbf{$Rec$}}
    & \multicolumn{1}{c}{\textbf{$T_{train} [s]$}}
    & \multicolumn{1}{c}{\textbf{$T_{pred} [s]$}} \\
            \midrule

\textit{Yahoo} &      &  &       &       &       &       &       &       &          &         \\
Arts      & \svm & 1.654 & \textbf{1.563} & 0.921 & 0.202 & 0.318 & 0.375 & 0.411 & 1258.310 & 12.868  \\
          & \nbc &       & 5.874 & 0.770 & 0.056 & 0.195 & 0.132 & \textbf{0.468} & 1820.328 & 276.032 \\
          & \nbx &       & 1.002 & \textbf{0.937} & \textbf{0.313} & \textbf{0.393} & \textbf{0.504} & 0.393 & 32.003   & 73.483  \\
          \mline
          & \tm  &       & -     & 0.930 & 0.355 & 0.442 & 0.483 & 0.483 & 31.814   & 57.090  \\
          \midrule
Business  & \svm & 1.599 & \textbf{1.586} & 0.970 & 0.516 & \textbf{0.702} & 0.812 & \textbf{0.802} & 1466.094 & 20.457  \\
          & \nbc &       & 3.484 & 0.902 & 0.181 & 0.408 & 0.307 & 0.669 & 2018.329 & 415.166 \\
          & \nbx &       & 1.012 & \textbf{0.971} & \textbf{0.541} & 0.682 & \textbf{0.868} & 0.682 & 44.473   & 119.584 \\
          \mline
          & \tm  &       & 1.599 & 0.977 & 0.761 & 0.810 & 0.832 & 0.832 & 44.468   & 91.686  \\
\midrule
Computers      & \svm & 1.507 & \textbf{1.289} & 0.959 & 0.400 & 0.500 & 0.575 & 0.482 & 2879.838   & 380.278  \\
               & \nbc &       &       &       &       &       &       &       &  &          \\
               & \nbx &       & 1.000 & \textbf{0.962} & \textbf{0.417} & \textbf{0.502} & \textbf{0.627} & \textbf{0.502} & 5.474      & 71.410   \\
               \mline
               & \tm  &       & -     & 0.955 & 0.434 & 0.505 & 0.539 & 0.539 & 5.462      & 47.888   \\
               \midrule
Education      & \svm & 1.463 & \textbf{1.262} & 0.949 & 0.228 & \textbf{0.326} & \textbf{0.416} & 0.360 & 2881.183 & 106.067  \\
               & \nbc &       & 7.009 & 0.787 & 0.054 & 0.182 & 0.102 & \textbf{0.492} & 2950.843 & 2403.130 \\
               & \nbx &       & 1.020 & \textbf{0.950} & \textbf{0.260} & 0.318 & 0.396 & 0.318 & 5.471    & 67.019   \\
               \mline
               & \tm  &       & -     & 0.946 & 0.282 & 0.346 & 0.377 & 0.377 & 5.763    & 42.000  \\
               \midrule
Entertainment  & \svm & 1.414 & \textbf{1.057} & \textbf{0.944} & 0.344 & \textbf{0.432} & \textbf{0.618} & 0.46  & 2494.289   & 70.982   \\
               & \nbc &       & 4.578 & 0.792 & 0.124 & 0.283 & 0.177 & \textbf{0.571} & 2473.593   & 1622.918 \\
               & \nbx &       & 1.000 & 0.927 & \textbf{0.353} & 0.393 & 0.449 & 0.393 & 5.195      & 46.373   \\
               \mline
               & \tm  &       & -     & 0.931 & 0.371 & 0.444 & 0.476 & 0.476 & 5.194      & 28.556   \\
               \midrule
Health         & \svm & 1.644 & \textbf{1.513} & \textbf{0.956} & \textbf{0.396} & \textbf{0.532} & \textbf{0.582} & 0.536 & 1256.259   & 76.788   \\
               & \nbc &       & 4.208 & 0.878 & 0.116 & 0.317 & 0.232 & \textbf{0.594} & 1781.850   & 1776.902 \\
               & \nbx &       & 1.065 & 0.953 & 0.338 & 0.439 & 0.565 & 0.446 & 3.689      & 45.598   \\
               \mline
               & \tm  &       & -     & 0.955 & 0.412 & 0.523 & 0.575 & 0.575 & 3.687      & 31.464   \\
               \midrule
Recreation     & \svm & 1.429 & \textbf{1.016} & \textbf{0.937} & 0.268 & 0.352 & \textbf{0.520} & 0.372 & 3118.926   & 77.415   \\
               & \nbc &       & 4.564 & 0.806 & 0.102 & 0.270 & 0.189 & \textbf{0.607} & 1442.366   & 1599.069 \\
               & \nbx &       & 1.003 & 0.936 & \textbf{0.373} & \textbf{0.436} & 0.517 & 0.436 & 4.870      & 45.549   \\
               \mline
               & \tm  &       & -     & 0.931 & 0.411 & 0.469 & 0.497 & 0.497 & 4.974      & 28.765   \\
               \midrule
Reference      & \svm & 1.174 & \textbf{1.087} & \textbf{0.966} & \textbf{0.423} & \textbf{0.488} & \textbf{0.527} & \textbf{0.488} & 1018.374   & 96.898   \\
               &      &       &       &       &       &       &       &       & &          \\
               & \nbx &       & 1.000 & 0.964 & 0.391 & 0.437 & 0.487 & 0.437 & 3.680      & 56.756   \\
               \mline
               & \tm  &       & -     & 0.962 & 0.418 & 0.449 & 0.464 & 0.464 & 3.807      & 29.016   \\
               \midrule
Science        & \svm & 1.450 & \textbf{1.253} & \textbf{0.957} & 0.227 & \textbf{0.312} & \textbf{0.388} & \textbf{0.334} & 2076.872   & 683.086  \\
               & \nbc &       &       &       &       &       &       &       & &          \\
               & \nbx &       & 1.000 & \textbf{0.957} & \textbf{0.243} & 0.297 & 0.375 & 0.297 & 3.234      & 52.596   \\
               \mline
               & \tm  &       & -     & 0.951 & 0.253 & 0.305 & 0.331 & 0.331 & 3.252      & 34.089   \\
               \midrule
Social         & \svm & 1.279 &       &  &  &       &       &       &           &          \\
               & \nbc &       &       &  &  &       &       &       &           &          \\
               & \nbx &       & \textbf{1.000} & \textbf{0.973} & \textbf{0.515} & \textbf{0.561} & \textbf{0.620} & \textbf{0.561} & 7.590      & 126.672  \\
               \mline
               & \tm  &       & -     & 0.969 & 0.551 & 0.578 & 0.592 & 0.592 & 7.569      & 71.701   \\
               \midrule
Society        & \svm & 1.670 & \textbf{1.425} & 0.935 & 0.281 & \textbf{0.415} & 0.474 & \textbf{0.404} & 10051.579  & 353.171  \\
               & \nbc &       &       &       &       &       &       &       &            &          \\
               & \nbc &       & 1.008 & \textbf{0.939} & \textbf{0.285} & {0.372} & \textbf{0.494} & 0.373 & 6.969      & 71.041   \\
               \mline
               & \tm  &       & -     & 0.934 & 0.314 & 0.406 & 0.450 & 0.450 & 6.928      & 52.596  \\
          \midrule
          \multicolumn{2}{r}{$Average\left(\Delta_{Loss}\right)$} & & & 0.089 & 0.068 & 0.041 & 0.056 & -0.037 & &\\
          \multicolumn{2}{c}{{ (\nbx \textit{vs} best opponent)}} & &  & & & & &  & &  \\          
          \bottomrule
         
\end{tabular}}
\caption{Metrics---Multinomial {BoW} Features (text data), ``CV 66\%'' applies.}
\label{tab:res_bow}
\end{table}

\begin{table}[]
\centering
{
\begin{tabular}{lrrrrr}
\toprule
{Data}& $\Delta H_{s}$ & $\Delta Z_{s}$& $\Delta Acc$ & $\Delta Pre$& $\Delta Rec$ \\
\midrule
Yeast         & -0.094  & -0.088  & -0.136   & -0.139   & -0.189   \\
Scene         & -0.040  & -0.189  & -0.071   & -0.121   & -0.070   \\
Emotions      & -0.035  & -0.045  & -0.064   & -0.043   & -0.123   \\
CAL500        & -0.046  & 0.002   & 0.005    & -0.231   & -0.190   \\
Mediamill     & -0.054  & -0.043  & -0.234   & -0.484   & -0.358   \\
Music         & -0.036  & -0.042  & -0.062   & -0.044   & -0.131   \\
NUS-WIDE      & -0.011  & -0.071  & 0.022    & 0.031    & -0.128   \\
\midrule
$Average\left(\Delta_{Loss}\right)$ & -0.045  & -0.068  & -0.077   & -0.147   & -0.170   \\
              &         &         &          &          &          \\
\midrule
Bibtex        & -0.001  & -0.014  & -0.070   & -0.027   & -0.379   \\
Enron         & -0.024  & -0.127  & -0.181   & -0.203   & -0.191   \\
LangLog          & 0.000   & 0.183   & 0.053    & -0.051   & -0.509   \\
SlashDot      & 0.004   & 0.011   & -0.002   & -0.094   & -0.037   \\
\midrule
$Average\left(\Delta_{Loss}\right)$ & -0.005  & 0.013   & -0.050   & -0.094   & -0.279   \\
              &         &         &          &          &          \\
\midrule
Arts          & 0.016   & 0.111   & 0.075    & 0.129    & -0.075   \\
Business      & 0.002   & 0.025   & -0.020   & 0.056    & -0.119   \\
Computers     & 0.003   & 0.017   & 0.002    & 0.052    & 0.020    \\
Education     & 0.001   & 0.032   & -0.008   & -0.020   & -0.174   \\
Entertainment & -0.017  & 0.009   & -0.039   & -0.169   & -0.178   \\
Health        & -0.003  & -0.058  & -0.093   & -0.017   & -0.148   \\
Recreation    & -0.001  & 0.105   & 0.084    & -0.003   & -0.171   \\
Reference     & -0.002  & -0.032  & -0.051   & -0.040   & -0.051   \\
Science       & 0.000   & 0.016   & -0.015   & -0.013   & -0.037   \\
Social        & 0.973   & 0.515   & 0.561    & 0.620    & 0.561    \\
Society       & 0.004   & 0.004   & -0.043   & 0.020    & -0.031   \\
\midrule   
$Average\left(\Delta_{Loss}\right)$ & 0.089   & 0.068   & 0.041    & 0.056    & -0.037  \\
\midrule

\end{tabular}}
\caption{\nbx \textit{vs} Best Opponent (positive values indicate \nbx is the best)}
\label{tab:deltas}
\end{table}


In the analysis of binary \textit{BoW} data (Table \ref{tab:bin_res_bow}) the differences between \nbx and the best options are relatively moderate with the exception of the \textit{Enron} data, where \nbx, despite capturing well the average $LCard$ yields scores inferior to those of other methods. In the case of \textit{LLog} data, 4 out of 5 methods predicted a cardinality of less than one, which is justified by the fact that 207 out of 1460 observations ($\sim$14\%) are not labeled (not excluded \textit{a priori}), 205 of which were correctly identified by \nbx.
On the other hand, for \textit{SlashDot}, there were three such cases although no observations had empty targets. These were two examples of weakly multi-labeled data sets that proved to be challenging to address.

For multinomial \textit{BoW} features (Table \ref{tab:res_bow}), \svm and \nbx underestimate the average cardinality, even though \nbx sensibly more than \svm. These two methods outperform \nbc and present scores very close to each other (see Table \ref{tab:deltas}), while differing by 3 to 4 orders of magnitude in terms of computing times for reasons already listed in Section \ref{sec:cont_results}.
For \nbc, both for continuous and \textit{BoW} features, we can observe a strong relation between the (over)estimated cardinality and a good score in recall, thanks to fact that this metric does not penalize including more labels than necessary in the target. As expected, different metrics capture different characteristics and they should be taken into consideration together when choosing a tool for a specific task.

\subsection{Prediction of the Target Size $m$}
\label{sec:truem}

The prediction of the parameter $m$ is of fundamental importance for the efficacy of our method and improvement on this alone would give great benefits. Intuitively, when we estimate too big or too small targets, we lose in terms of performance, as we are making mistakes no matter the result of the subsequent estimations.
For continuous features, \nbx, \svm and \esvm generally manage to capture well the cardinality of the data sets while for text data they underestimate the size of the targets, especially in the case of \nbx. \nbc and \enbc tend in general to overestimate the size of targets, leaving space for a case-by-case parameter tuning to possibly get better results.
On most of the instances, \tm is as good as or better than the best performing algorithm on all metrics, prompting great interest for further work on the specific topic of size estimation.
This, though not a trivial task, can be considered as a separate step from label prediction, for example in the form of a standard multi-class classification problem. Numerous appropriate solutions already exist for this problem, whose properties are well studied and for which substantial literature is available \citep{hastie2001}. With our experiments, we could show how much \nbx can gain from improvements in this single step alone. In cases where we have a large label set for example, the time spent on \texttt{predict\_m} is a small fraction of that spent on \texttt{predict\_y}. Adopting more sophisticated methods at the cost of increased computational time is an option worth exploring.

\section{Conclusion}
\label{sec:conclusion}
This paper showed how a MLC problem can be approached via a cascade of predictors and how to employ NBC to carry out prediction, yielding an extension of Naive Bayes Classification to the domain of multi-label classification. The proposed algorithm showed significant advantages in terms of computation costs and proved to be competitive in terms of predictive performance, thus offering a viable alternative for tasks requiring a more agile computational footprint.

Our approach allows to see the problem from a different perspective than the current literature, notably thanks to the prediction of the target size ($m$) independently from the prediction of the labels. In future research we will further address this aspect, not excluding the possibility of mixing different prediction paradigms for the two tasks.

In its current formulation, \nbx does not guarantee that the solution found is optimal. To compensate for this, an interesting research lead consists in the possibility of predicting more than one target vector at a time, returning as final answer a set of vectors with their relative score. An interesting feature of \nbx is that this prediction can be done with little added complexity. Additionally, the very low computational footprint of \nbx opens the possibility of applying a committee-learning scheme (Boosting for instance \citep{hastie2001}) to improve the overall performance. 

Overall, we introduced \nbx, a computationally light and efficient multi-label classification method, that proved to be both scalable to large and complex data sets and competitive with state-of-the-art algorithms in terms of predictive performance. This opens up new perspectives for its application on large scale, real-world data.

\section*{Acknowledgements}
This research benefited from the support of the ``FMJH Program Gaspard Monge in optimization and operation research'', and from the support to this program from EDF.

\bibliography{naibx}

\end{document}